%% file: main.tex
\documentclass[twocolumn]{bmcart}

\usepackage[utf8]{inputenc} 

\usepackage[section]{placeins}
\usepackage{multirow}
\usepackage{xcolor}
\usepackage{multirow}
\usepackage{graphicx}
\usepackage{subcaption}

\usepackage{booktabs}
\usepackage{fancyvrb}
\usepackage{listings}
\usepackage{float}

\usepackage{tikzducks}
\usepackage[inline]{showlabels}

\usepackage[draft,inline,nomargin,index]{fixme}
\fxsetup{theme=color,mode=multiuser}
\FXRegisterAuthor{ft}{envft}{\color{red}FT}
\FXRegisterAuthor{ag}{envag}{\color{blue}AG}
\fxsetup{layout={footnote}}
\fxusetheme{colorsig}

\usepackage{booktabs,ragged2e}

\usepackage{algpseudocode}
\usepackage{algorithmicx}
\usepackage{algorithm}
\usepackage{algpseudocode}

\usepackage{natbib}
\usepackage{hyperref}
\usepackage[flushleft]{threeparttable}

\usepackage{graphicx}
\usepackage{balance}
\DeclareUnicodeCharacter{2212}{-}
\usepackage{multicol}
\usepackage{multirow}
\usepackage{multicol}

\usepackage{booktabs}

\usepackage{amsmath}
\newcommand\numberthis{\addtocounter{equation}{1}\tag{\theequation}}

\usepackage[hypcap=false]{caption}

\usepackage{amsthm,amsmath,amsfonts}
\usepackage[utf8]{inputenc} 

\startlocaldefs
\endlocaldefs

\begin{document}

\begin{frontmatter}

\begin{fmbox}
\dochead{Research}

\title{Privacy-preserving Logistic Regression with Secret Sharing}
\author[
   addressref={aff1},                   
   corref={aff1},                       
   email={a.r.ghavamipour@rug.nl}   
]{\inits{ARG}\fnm{Ali Reza} \snm{Ghavamipour}}
\author[
   addressref={aff1},
   email={f.turkmen@rug.nl}
]{\inits{FT}\fnm{Fatih} \snm{Turkmen}}
\author[
   addressref={aff2},
   email={email}
]{\inits{XJ}\fnm{Xiaoqian} \snm{Jiang}}


\address[id=aff1]{
  \orgname{University of Groningen}, 
  \street{Nijenborgh 9},                     %
  \city{Groningen},                              
  \cny{Netherlands}                                    
}
\address[id=aff2]{%
  \orgname{UTHealth School of Biomedical Informatics, The University of Texas},
  \city{Houston},
  \cny{US}
}








\begin{abstractbox}

\begin{abstract} 

\parttitle{Background} Logistic regression (LR) is a widely used classification method for modeling binary outcomes in many medical data classification tasks. Research that collects and combines datasets from various data custodians and jurisdictions can excessively benefit from the increased statistical power to support their analyzing goals. However, combining data from these various sources creates significant privacy concerns that need to be addressed. 

\parttitle{Methods} In this paper, we proposed secret sharing-based privacy-preserving logistic regression protocols using the Newton-Raphson method. Our proposed approaches are based on secure Multi-Party Computation (MPC) with different security settings to analyze data owned by several data holders. 

\parttitle{Results} We conducted experiments on both synthetic data and real-world datasets and compared the efficiency and accuracy of them with those of an ordinary logistic regression model. Experimental results demonstrate that the proposed protocols are highly efficient and accurate.

\parttitle{Conclusions} This study introduces iterative algorithms to simplify the federated training a logistic regression model in a privacy-preserving manner. Our implementation results show that our improved method can handle large datasets used in securely training a logistic regression from multiple sources.

\end{abstract}




\end{abstractbox}
\end{fmbox}

\end{frontmatter}


\input{intro}

\input{related}
\input{preliminaries}
\input{approach}

\input{results}

\input{conclusions}

\begin{backmatter}

\section*{Declarations}

\section*{Abbreviations}

LR, logistic regression; MPC, multi-party computation; GLORE, grid binary logistic regression;SMAC-GLORE, Secure Multi-pArty Computation Grid LOgistic REgression; BMPC, beaver triple-based multi-party computation; CMPC, classical multi-party computation; OLR, ordinary logistic regression; PIMA, pima indians diabetes dataset; LBW, low birth weight study; PCS, prostate cancerstudy; UIS, umaru impact study datasets

\section*{Availability of data and materials}

\href{https://github.com/alirezaghavamipour/pplr_ss}{github.com/alirezaghavamipour/pplr\_ss}
\section*{Authors’ contributions}

The author ARG contributed the majority of the writing and conducted major parts of the study. FT wrote a part of this paper, provided detailed edits and critical suggestions. XJ provided the motivation for this work and helpful comments. All authors read and approved the final manuscript.

\section*{Competing interests}

The authors declare that they have no competing interests.

\section*{Consent for publication}

Not applicable.

\section*{Ethics approval and consent to participate}

Not applicable.

\section*{Funding}

Not applicable.

\section*{Acknowledgment}

We thank Aida Plocco and Sytze Tempel for their contribution to the idea of this study.

\bibliographystyle{bmc-mathphys} 
\bibliography{biblio}      

\end{backmatter}
\end{document}

%% file: intro.tex
\section{Background}

Nowadays, patient data (i.e., medical records and genomics) are rapidly being collected worldwide, which has expanded in volume faster than anyone predicted. Medical data analysis often benefits from collecting more samples to increase the power of statistical analysis and the robustness of machine learning models. To acquire more valuable information for research, hospitals and research centers would like to collaborate by sharing their data and findings in a central location. The main benefits of the collaborative sharing and processing of data are more accurate detection and diagnosis of disorders, prediction of disease origin, and development of drugs. 

Various data analytic techniques can be employed to extract information from the shared datasets. The logistic regression model \cite{hosmer2013applied}, one of the most popular prediction models, is now widely used in medical research. By analyzing data, logistic regression estimates a particular event's probability based on previously observed data and creates a binary classification model. More specifically, the value of a binary variable is predicted based on several independent variables. For example, logistic regression could develop a model suitable for identifying malignant breast cancers based on tumor size, patient age, blood type, and genetic inheritance \cite{boxwala2011using}. Statistical models need a sufficient sample size to ensure necessary statistical power in data analysis, and machine learning models require even a large sample size (due to the consideration of interactions and non-linearity assumptions) \cite{riley2020calculating}. It is also beneficial to combine and compare data from different sources (i.e., ensuring generalizability). However, collecting data from multiple sources often lead to privacy concerns. Due to institutional policies and legal restrictions, hospitals and medical centers often have concerns about sharing their sensitive data (i.e., especially patient-level information) with other institutions. Therefore, it is essential to use privacy-preserving solutions and techniques, such as secure Multi-Party Computation (MPC), to train a logistic regression model over shared data between multiple data holders \cite{jagadeesh2017deriving}.

In this paper, our particular focus is on enabling logistic regression between multiple data holders, such as hospitals and research centers, in a privacy-preserving manner. With the proposed protocols, multiple parties can train a logistic regression model using the Newton-Raphson optimization method under different security assumptions. Thus, the main contributions of this paper include:
\begin{itemize}
    \item A novel privacy-preserving algorithm for computing logistic regression models that is very accurate but not as efficient, although acceptable.
    \item A second algorithm that is very efficient but less accurate due to the use of multiple approximations.
    \item Implementation of the proposed algorithms in both honest and dishonest majority security settings.
    \item Evaluation of the proposed protocols on various real-world and generated synthetic datasets.
\end{itemize}

We consider a setting in which data are horizontally partitioned, which indicates that data holders have precisely the same variables but different values for those variables.

%% file: related.tex
\section{Related Work}
Various approaches make use of cryptographic techniques such as multi-party computation (MPC), homomorphic encryption, and differential privacy. Previous studies showed the practicability of building a secure distributed logistic regression across multiple data holders. However, due to the complexity of the underlying secure computation primitives, their methods suffer from multiple issues such as scalability, accuracy loss, and low efficiency.

The Grid Binary LOgistic REgression (GLORE) model was proposed by Wu et al. \cite{shi2016secure} to support privacy-preserving logistic regression in a distributed manner. GLORE estimates global model parameters for horizontally partitioned data without necessarily sharing patients. Rather than directly sharing sensitive data with other institutes, the decomposable intermediary results with significantly less sensitive information are transferred to build a global training protocol for logistic regression. However, in their proposed methods, sensitive data could leak due to disclosure of the information matrix and summary statistics.

Shi et al. \cite{shi2016secure} proposed a secure multi-party computation framework for grid logistic regression (SMAC-GLORE), which protects the patient data's confidentiality and privacy. SMAC-GLORE preserves the intermediary results based on garbled circuits during iterative model learning. Various approaches such as secure matrix multiplication and addition and fixed-Hessian methods have been employed to estimate the maximum likelihood. However, due to the garbled circuit constraints, SMAC-GLORE cannot handle a large number of records and features. Also, it uses the Taylor series approximation approach to evaluate the sigmoid function, which causes accuracy loss in the final result.


Xie et al. \cite{xie2016privlogit} developed PrivLogit, which performs distributed privacy-preserving logistic regression and uses Yao's garbled circuits and Paillier encryption. PrivLogit needs the data owners to perform computations on their data before encryption to compute parts of a logistic regression matrix. However, their proposed method requires an expensive computational cost to calculate the intermediate results.

Cock et al.\cite{pop00001} proposed an information-theoretically privacy-preserving model training protocols based on secret sharing-based building blocks such as distributed multiplication, distributed comparison, bit-decomposition of shares. Similar SecureML, their proposed protocol requires multiplication triples distributed during a setup phase with or without a trusted authority. Unlike SecureML, which is secure in the computational context, they engage in the information-theoretic model using secret sharing-based MPC and employ commodity-based cryptography \cite{beaver1997commodity} to decrease the number of communications.

SecureML~\cite{mohassel2017secureml} was one of the fastest protocols for privacy-preserving logistic regression models training based on secure MPC. The SecureML protocol is divided into an offline (to generate and distributing multiplication triples) and an online phase. Their proposed multiplication protocol is based on a straightforward and efficient security setting introduced by Beaver \cite{beaver1997commodity}. Also, to compute the activation functions, they proposed a new comparison-based activation function that converges to 0 and 1. Unlike our work that employs the Newton Raphson optimization method, SecureML focuses on the mini-batch gradient descent.

Other than MPC-based solutions, two popular methods have been considered. The first one is homomorphic encryption \cite{gentry2009fully}, which allows for computation to be performed over encrypted data, and has been applied to privacy-preserving logistic regression \cite{pop00026}\cite{pop00007}\cite{pop00015}\cite{pop00011}\cite{pop00020}\cite{pop00021}\cite{pop00008}. In most of these works, polynomial approximations need to be made to evaluate non-linear functions in machine-learning algorithms. The second method is differential privacy, a universally accepted mathematical structure for protecting data privacy. The main application of differential privacy in machine learning is when the model is published publicly after training in a way that personal data points cannot be distinguished from the released model \cite{pop00019}\cite{chaudhuri2009privacy}\cite{el2013secure}\cite{kim2019secure}.

%% file: preliminaries.tex
\section{Preliminaries}\label{preliminaries}

\subsection{Secure multi-party computation}

Secure Multi-Party Computation (MPC) allows computation parties to compute an arbitrarily agreed function of their private inputs. During the computation, no party should reveal its private inputs to the other parties or any third party. This is formalized as a secure function evaluation where n parties compute a function

\begin{equation}
    f(x_1,x_2,...,x_n) = (y_1,y_2,...,y_n) 
\end{equation}

such that each computation party $P_i$, where $1 \leq i \leq n $, provides its input $x_i$ and learns its output $y_i$. The secret shared value can be exposed to each party by combining with other parties shares. Since our MPC solution is based on additive secret sharing, we briefly discuss this concept in the next section.

\subsection{Secret sharing}

Secret sharing is a set of techniques that allows a secret value $x$ to be distributed among $n$ participants as $x_1, . . . , x_n$ so that each party $P_i$ receives a random share $x_i$ $(mod ~p)$ over some prime $p$ \footnote{the modular notation is dropped for the means of conciseness and ease of composition}. In secret sharing-based secure computation schemes, a different number of sensitive data holders (input parties) can secretly share their data among other participants. In this paper, we use the n-out-of-n additive secret sharing scheme. In this scheme, an integer $u$ additively shares between $n$ participants. In other words, each input party pick $n - 1$ randomly generated values and sends them to all other participants. Also, one of the parties is provided by the secret $u$ minus the sum of those randomly generated values, which permits the reconstruction of the original value by summing all of the shares. 

In what follows, we will use $[\![ x ]\!]$ to denote secret shares that reconstruct to $x$. A share $[\![ x ]\!]$ is an n-tuple with each computing party holding precisely one element of the tuple and $[\![ x ]\!]_{i}$ denotes the share held by the $i_{th}$ party.



\subsection{Security model} \label{security}

We consider a set of input parties who aim to train a logistic regression model on their sensitive data. We assume the data is distributed horizontally among the input parties, where each independent database contains only a sub-population. Furthermore, we consider a set of computation parties $n$ that receives secret-shared data in a setup phase (this phase is not considered in any future computation). 

We assume the computation parties are non-colluding (the servers cannot be controlled by one authority), independent (if the adversary controls one party, other parties behave honestly), and honest but curious (each party correctly follows the protocol but might be curious about the information transferred between other parties). However, one or more computation parties may get corrupted, and these corrupted parties could involve more than half of the computation parties. We propose logistic regression training protocols for both honest majority and dishonest majority assumptions in this paper.  In the case with the honest majority assumption, the adversary may actively corrupt $t < n/2$ players. We implement this protocol in three-party settings where only one party can be corrupted at most. In the second protocol, we propose a protocol in which the number of corrupted parties could be more than $n/2$. To achieve the highest efficiency, we implement this protocol in the two-party setting.

\subsection{Addition and multiplication}
Various operations can be performed on secret shared data through tailored protocols. Based on the considered security models mentioned in section \ref{security}, we employ addition and multiplication as the key operations in our work. Under both introduced security models, the addition of two secrets can be performed locally without any communication: $[\![ x ]\!] + [\![ y ]\!] = ([\![ x ]\!]_1 + [\![ y ]\!]_1, [\![ x ]\!]_2 + [\![ y ]\!]_2,\ldots , [\![ x ]\!]_n + [\![ y ]\!]_n)$. However, the multiplication of additively secret shared values requires network communication. 

In the honest majority setting, we use the multiplication protocol proposed by Bogdanov et al. \cite{bogdanov2012high} (classical approach). Their protocol is based on the following equation:

        \begin{align*}
            xy = (x_1 +x_2 +x_3)(y_1 +y_2 +y_3) = \\
            (x_1y_1 + x_1y_3 + x_3y_1) +  \\
            (x_2y_2 + x_2y_1 + x_1y_2) +\\
            (x_3y_3+x_3y_2+x_2y_3)\\
            = \sum_{i=1}^{3}\sum_{j=1}^{3}x_{i}y_{j}
            \numberthis \label{mulg}
        \end{align*}

As shown in Equation \ref{mulg}, to perform the multiplication operation, each computation party requires its adjacent computation party's input share. However, Bogdanov et al. stated that sending the input share to the nearby computation party may reveal more information about the input data than necessary. To solve this issue, they introduced a re-sharing protocol to distribute a new share of the original value in each computation party at the beginning and end of the multiplication operation. 

In the dishonest majority setting, we use the Beaver triples technique \cite{beaver1991efficient} to perform multiplication. This method requires a trusted initializer to pre-distribute the multiplication triples (random and independent) shares ($[\![ a ]\!]$, $[\![ b ]\!]$, $[\![ c ]\!]$) between the computation parties in a way that $a . b = c$. To perform the multiplication, each computation party computes $[\![ d ]\!]$ = $[\![ x ]\!]$ - $[\![ a ]\!]$ and $[\![ e ]\!]$ = $[\![ y ]\!]$ - $[\![ b ]\!]$ locally and then reveals both $[\![ d ]\!]$ and $[\![ e ]\!]$. Revealing these shares does not compromise the security of sensitive data, as $[\![ a ]\!]$ and $[\![ b ]\!]$ have been randomly generated and thus mask the secret values. Next, each party locally computes: 

\begin{align*}
[\![ w ]\!]_i = [\![ c ]\!]_i +[\![ e ]\!].[\![ b ]\!]_i +[\![ d ]\!].[\![ a ]\!]_i +[\![ e ]\!].[\![ d ]\!] \\
\end{align*}

where $[\![ w ]\!]_i$ is a share of the result of the multiplication. 

Note that after distributing the multiplicative triples needed, the trusted initializer is not involved in the rest of the protocol. Thus, the trusted initializer does not understand the function to be computed or the computation inputs.

\subsection{Inversion} \label{inv}
 As we will see in Section~\ref{sec:approach}, we will need the operations of inversion and matrix inversion in order to implement logistic regression. The MPCs protocols discussed previously support only addition and multiplication. Since the accurate implementation of inversion significantly increases the computational cost, we use the approximation method introduced by Nardi~\cite{nardi2012achieving}. Nardi's method converts the matrix inversion problem into an iterative procedure of matrix multiplication and addition. In this method, we look for a matrix $X$ to find the inversion of the matrix $B$ according to:

        \begin{equation}
            X^{-1} = B \nonumber
        \end{equation}

The main idea is to define a function of which matrix $X$ represents its root $(f(X)=0)$. Therefore, the function $f(x)$ is defined as follows:

\begin{equation}
f = X^{-1} - B \nonumber
\end{equation} 

To find the root of the function $f$ , Nardi suggested using the Newton-Raphson method.  By applying this method (take the derivative of function $f(X)$ and applying the general iterative Newton method), a stable numerical iterative approximation takes the following form:

            \begin{align}
                    X_{s+1} = 2X_s - X_sM_s \quad X_0 = c^{-1}\mathbb{I}  \\
                    M_{s+1} = 2M_s - M_s^2 \quad  M_0 = c^{-1}B  \nonumber
            \end{align} 

where $X_0$ and $M_0$ are the initial guesses, $\mathbb{I}$ is an identity matrix, and $c$ is a constant. After convergence, $X_{s}$ contains an approximation of matrix $B$'s inversion.

\subsection{Logistic regression}

Logistic regression is a standard machine learning technique that is commonly used in various areas of research. It predicts the probability that a dependent variable belongs to a class. This paper will consider the binary classification, where the dependent variable belongs to two possible classes. The logistic model is intended to describe a probability, which is always a number between 0 and 1. 

Assume a training dataset $D = \{(x_1, y_1), (x_2, y_2)$ $,\dots, (x_n, y_n)\}$ of n records, where $x_i$ is the m-dimensional feature vector of each record and the $y_i$ is a vector of labeled binary outcomes. The logistic regression model is given by:

            \begin{equation}
                P(y_i=1|x;\beta) = \pi(\beta^Tx_i)= {\frac{1}{1-e^{-\beta^Tx_i}}} 
            \end{equation}

where $\beta = (\beta_0,\ldots, \beta_d)$ is p-dimensional regression coefficients vector, y is the observation of binary responses, and x is the feature vectors. The purpose of using this method is to obtain the parameter vector $\beta$ that maximizes the log-likelihood function: 
            \begin{equation}
                l(\beta) = -\sum^{n}_{i=1}\log(1+e^{-\beta^Tx_i})
            \end{equation}
            
By determining the parameters $\beta$, the classifier can predict the class label of new feature vectors.


            

%% file: approach.tex
\section{Methods}\label{sec:approach}
\subsection{Estimating model coefficients}

Since logistic regression cannot be found in a closed form, model estimation is often accomplished by an iterative optimization over the log-likelihood function. Newton-Raphson \cite{agresti2003categorical} is a numerical iterative method that eventually approaches the optimal values of the $\beta$ coefficients. For each iteration, the coefficient estimates are updated by:

            \begin{equation}\label{main_newton}
                \beta_{new} = \beta_{old} - \mathbb{H}^{-1}(\beta_{old}) \nabla(\beta_{old})
            \end{equation}

where $\nabla$ and $\mathbb{H}$ respectively correspond to the gradient and the Hessian of the log-likelihood function evaluated the current estimate of the $\beta$ coefficients. The gradient and Hessian for logistic regression can be computed as follows:

            \begin{equation}
                \nabla(\beta) = \frac{\partial l(\beta)}{\partial \beta} = \mathbb{X}^\mathbb{T}(y-\pi)
                \label{grad}
            \end{equation} 
            
            \begin{equation}
                \mathbb{H}(\beta) = \frac{\partial^2 f}{\partial \beta \partial \beta^\mathbb{T}} = \mathbb{X}^\mathbb{T}\mathbb{W}\mathbb{X} \label{hess}
            \end{equation}

where  $\mathbb{W}$ is a diagonal matrix with elements defined as $a_{i,i} = \pi(1 − \pi)$ and $\pi$ is the vector of probabilities.

\subsection{Gradient}\label{gradient}
To compute the gradient \eqref{grad}, first we need to compute the Sigmoid function ($\pi$). To do this, we consider two different cases in our paper:

            \begin{equation}
                \pi (z) =  \frac{\mathrm{1} }{\mathrm{1} + e^{-z}}  = (\mathrm{1} + e^{-z})^{-1}
                \label{sig}
            \end{equation}
            
a) Computation of the exact value of the Sigmoid function: The main challenges of computing the exact value of the Sigmoid function vector are the performing of exponentiation, inversion, and matrix inversion. To perform inversion and matrix inversion, we use the solutions discussed in Section \ref{inv}. However, performing exponentiation by the considered secret sharing techniques is still challenging. 

To tackle this issue, we implement exponentiation by using addition and multiplication. The solution is based on the idea of using additive secret sharing one more time. In other words, each computation party plays the role of a sensitive data holder. To do this, each computation party $i$ computes $e^{[\![ z_i ]\!]}$ locally and then computes $n$ different shares of it ($n$ indicates the total number of computation parties) and sends each of these values to the correspondence computation parties. After each computation party received the other computation parties' share of $e^{[\![ z_i ]\!]}$, by using the MPC multiplication operation, it computes $({{[\![e^{[\![ z_1 ]\!]}]\!]}_1} * {{[\![e^{[\![ z_2 ]\!]}]\!]}_2} * \dots *{{[\![e^{[\![ z_i ]\!]}]\!]}_n}$) which is equal to $([\![{e^{[\![ z_1 ]\!] + {[\![ z_2 ]\!]} + \dots + {[\![ z_i ]\!]}}}]\!])$. Therefore, each computation party has a valid share of $[\![e^{[\![ z ]\!]}]\!]$ and using the addition and inversion operation, the exact value of the Sigmoid function will be computed.



b) Least Squares Approximation of the Sigmoid Function: The introduced method to compute the sigmoid function's exact value might have scalability issues due to the substantial required number of multiplications. In order to improve the performance, we use the least-squares approximation of the sigmoid function over the interval [{{-8},{8}}] introduced by Kim et al. \cite{kim2018secure}. We adapt this approximation method and consider the degree 3, 5, and 7 least-squares polynomials:

\[
\begin{cases}
g_3(x) = 0.5 + 1.20096 . (x/8) - 0.81562 . (x/8)^3 \\\nonumber

g_5(x) = 0.5 + 1.53048 . (x/8) - 2.3533056 . (x/8)^3 \\\nonumber
\qquad\qquad + 1.3511295 . (x/8)^5
\\

g_7(x) = 0.5 + 1.73496 . (x/8) - 4.19407 . (x/8)^3 \\
\qquad\qquad + 5.43402 . (x/8)^5 - 2.50739 . (x/8)^7

\end{cases}
\]

The degree 3 least-squares approximation requires fewer multiplications, while the degree 7 polynomial has more immeasurable precision. 

\subsection{Hessian}

The Hessian matrix $\mathbb{H}$ denotes the second partial derivatives of the maximum likelihood function $l(\beta)$. In every iteration, the Hessian matrix has to be updated by the newest $\beta$, and its inversion has to be computed. To evaluate the Hessian matrix, we can consider two different methods. First, we can compute the exact value of the Hessian matrix by performing the required MPC-based multiplication. However, the exact evaluation of the Hessian matrix is considerably expensive in computational terms. To solve this issue, we approximate the Hessian matrix with a fixed matrix instead of updating it in every iteration. More specifically, we can replace the fixed Hessian matrix with its approximation $\Tilde{\mathbb{H}}$ (Equation \ref{hes}) that only needs to be computed and inverted once. 
\begin{equation}
    \Tilde{\mathbb{H}} = \frac{-1}{4}XX^T
    \label{hes}
\end{equation}

Böhning~\cite{bohning1999lower} has proved that if $\Tilde{\mathbb{H}} − \mathbb{H}$ is positive definite and $\Tilde{\mathbb{H}} \leq H$ then the convergence of this method is guaranteed. Also, because $\Tilde{\mathbb{H}}$ does not depend on $\beta$, we can pre-compute the Hessian and its inverse one time and used it in all iterations.

\subsection{Privacy-preserving Logistic Regression Training}

In this work, we assumed that the result party desires to compute the logistic regression model over collected data by different data owners. Each data owner computes multiple shares (based on the number of computation parties) of its sensitive data and separately sends them to each computation party. Note that each computation party receives an equal number of dependent $X_i$ and independent $y_i$ variables. Each computation party should append the received shares and their corresponding dependent variables in the correct order. Finally, computation parties send their computed shares of logistic regression coefficient to the result party, and the result party simply sums these shares together to compute the final result.

We now present our privacy-preserving logistic regression training algorithms that employ the previously mentioned approaches. These algorithms summarize the crucial steps in the proposed protocols for both honest and dishonest majority security assumptions. In our proposed algorithms, each data owner provides a share of data for the computation parties as input. The only output of the algorithm is the computed model coefficients $\beta$. Also, to prevent unnecessarily revealing information about the input, we will not employ a convergence check after each iteration. $n_{iter}$ specifies the upper bound of the number of iterations needed for convergence.

\begin{algorithm}
\caption{Accurate Logistic Regression Training algorithm}\label{pplr_accurate}
\begin{flushleft}
\hspace*{\algorithmicindent} \textbf{Input:} A share of input data ($[\![ X ]\!]$ ,  $[\![ y ]\!]$) from data owners, $n_{iter}$ number of iterations\\
\hspace*{\algorithmicindent} \textbf{Output} A share of computed vector of coefficient $[\![ \beta ]\!]$\\
\end{flushleft} 
\begin{algorithmic}[1]
\Procedure{}{}
\State $[\![ \beta ]\!] , [\![ \beta_{old} ]\!] \gets 0$
\State $\Tilde{\mathbb{[\![ H} ]\!]}\gets \frac{-1}{4}[\![ X ]\!] .  [\![ X ]\!]^T $
\State $\Tilde{\mathbb{[\![H ]\!]}} \gets {\Tilde{\mathbb{[\![H ]\!]}}}^{-1} $
\For {0 to $n_{iter}$}
\State $[\![ \beta ]\!]= [\![ \beta_{old}]\!]$
\State $ z \gets [\![ X ]\!] . [\![ \beta ]\!]$
\Procedure{Computing $[\![\pi]\!]$}{}
\State locally computes $e^{[\![ z ]\!]}$
\State send/receive a share of $e^{[\![ z ]\!]}$ to/from other CPs
\State $[\![e^{z}]\!]= [\![e^{[\![ z ]\!]}]\!]_1 . [\![e^{[\![ z ]\!]}]\!]_2...[\![e^{[\![ z ]\!]}]\!]_i$
\State $[\![{\pi}]\!] = [\![(1 + e^{z})]\!]^{-1} $
\EndProcedure
\State $[\![ \nabla(\beta_{temp})]\!] = [\![ y ]\!] - [\![ \pi ]\!]$
\State $[\![ \nabla(\beta)]\!] = [\![ X ]\!]^T  . [\![ \nabla(\beta_{temp})]\!]$
\State $[\![ \beta_{old}]\!] = [\![ \beta ]\!]  - [\![ {\Tilde{\mathbb{H}}^{-1}}]\!]  . [\![\nabla(\beta)]\!]$
\EndFor 
\State Return $[\![ \beta ]\!]$
\EndProcedure
\end{algorithmic}
\end{algorithm}

In Algorithm \ref{pplr_accurate}, we propose a very accurate privacy-preserving logistic regression model training protocol. In this algorithm, we only employ highly accurate approximations such as matrix inversion and fixed Hessian, which have a negligible effect on the computation output's accuracy. Moreover, instead of approximating the Sigmoid function, we use our introduced approach in section \ref{gradient} to compute the Sigmoid function's exact value (lines 8-12).

The main purpose of proposing Algorithm \ref{pplr} is to achieve a highly efficient privacy-preserving logistic regression model training protocol. Various approximation approaches such as fixed Hessian matrix, least-square approximation for the Sigmoid function, and matrix inversion algorithm are employed to obtain our goal. This logistic regression training algorithm demonstrates how the introduced approximation approaches can be efficiently combined to compute the logistic regression coefficient in a privacy-preserving manner.

\begin{algorithm}
\caption{Approximation-based Logistic Regression Training algorithm}\label{pplr}
\begin{flushleft}
\hspace*{\algorithmicindent} \textbf{Input:} A share of input data ($[\![ X ]\!]$ ,  $[\![ y ]\!]$) from data owners, $n_{iter}$ number of iterations\\
\hspace*{\algorithmicindent} \textbf{Output} A share of computed vector of coefficient $[\![ \beta ]\!]$\\
\end{flushleft} 
\begin{algorithmic}[1]
\Procedure{}{}
\State $[\![ \beta ]\!] , [\![ \beta_{old} ]\!] \gets 0$
\State $[\![ \Tilde{\mathbb{H}}]\!] \gets \frac{-1}{4}[\![ X ]\!] .  [\![ X ]\!]^T $
\State $[\![ \Tilde{\mathbb{H}}]\!] \gets [\![ {\Tilde{\mathbb{H}}]\!]^{-1}} $
\For {0 to $n_{iter}$}
\State $[\![\beta]\!] = [\![\beta_{old}]\!]$
\State $\textit{Compute}$ $[\![ \pi ]\!]$
\State $[\![\nabla(\beta_{temp}) = [\![ y ]\!] - [\![ \pi ]\!]$
\State $[\![\nabla(\beta)]\!] = [\![ X ]\!]^T  . [\![\nabla(\beta_{temp})]\!]$
\State $[\![\beta_{old}]\!] = [\![\beta]\!]  - [\![{\Tilde{\mathbb{H}}^{-1}}]\!]  . [\![\nabla(\beta)]\!]$

\EndFor 
\State Return $[\![ \beta ]\!]$
\EndProcedure
\end{algorithmic}
\end{algorithm}

%% file: results.tex
\section{Results}

\begin{figure*}
    \centering
    \begin{minipage}[c]{0.45\textwidth}
        \centering
        \includegraphics[width=\linewidth]{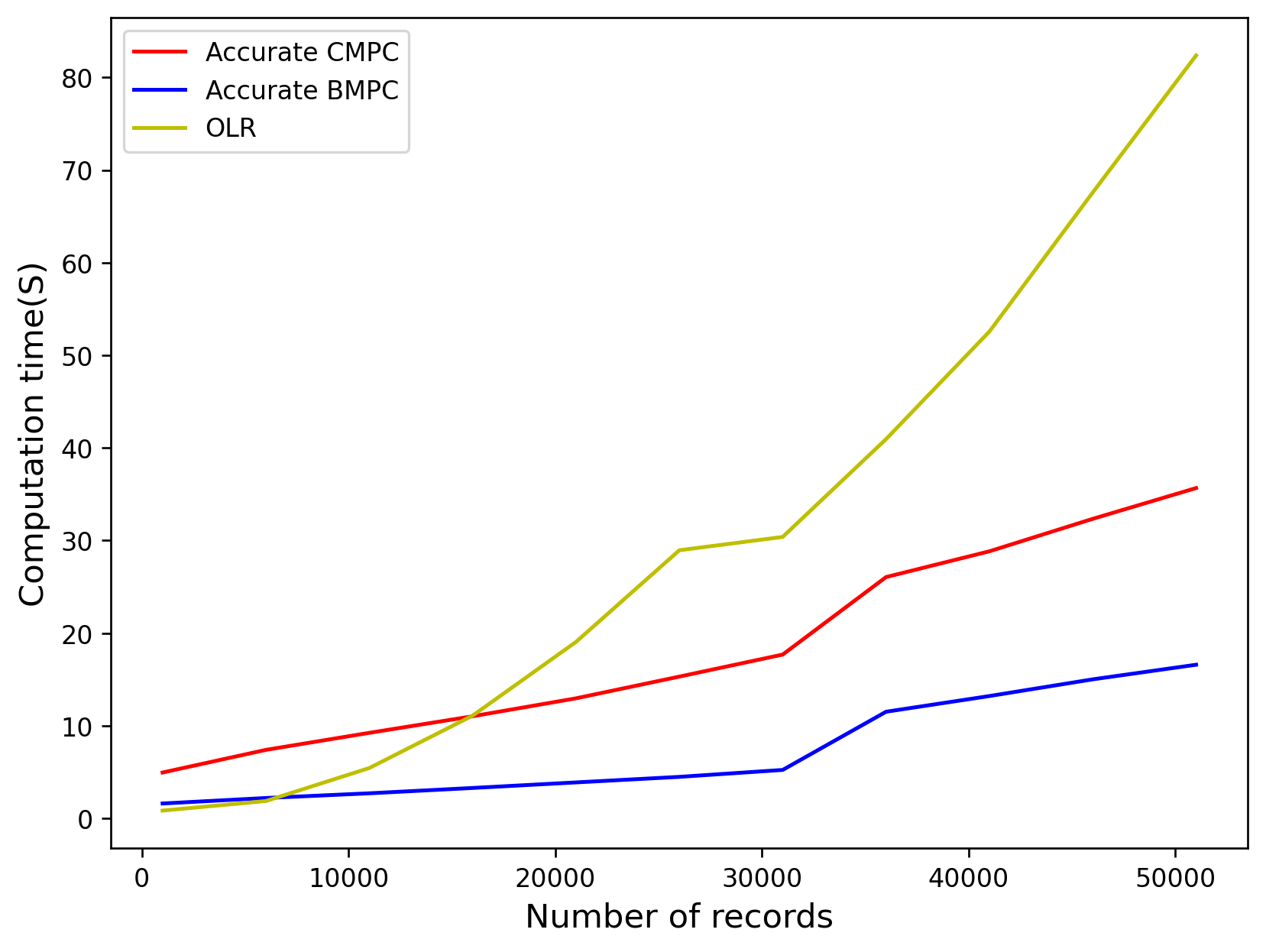}
        \caption{Efficiency comparison for increasing\\
        number of records using accurate algorithm  \ref{pplr_accurate}}
        \label{fig:1}
    \end{minipage}%
    \begin{minipage}[c]{0.45\textwidth}
        \centering
        \includegraphics[width=\linewidth]{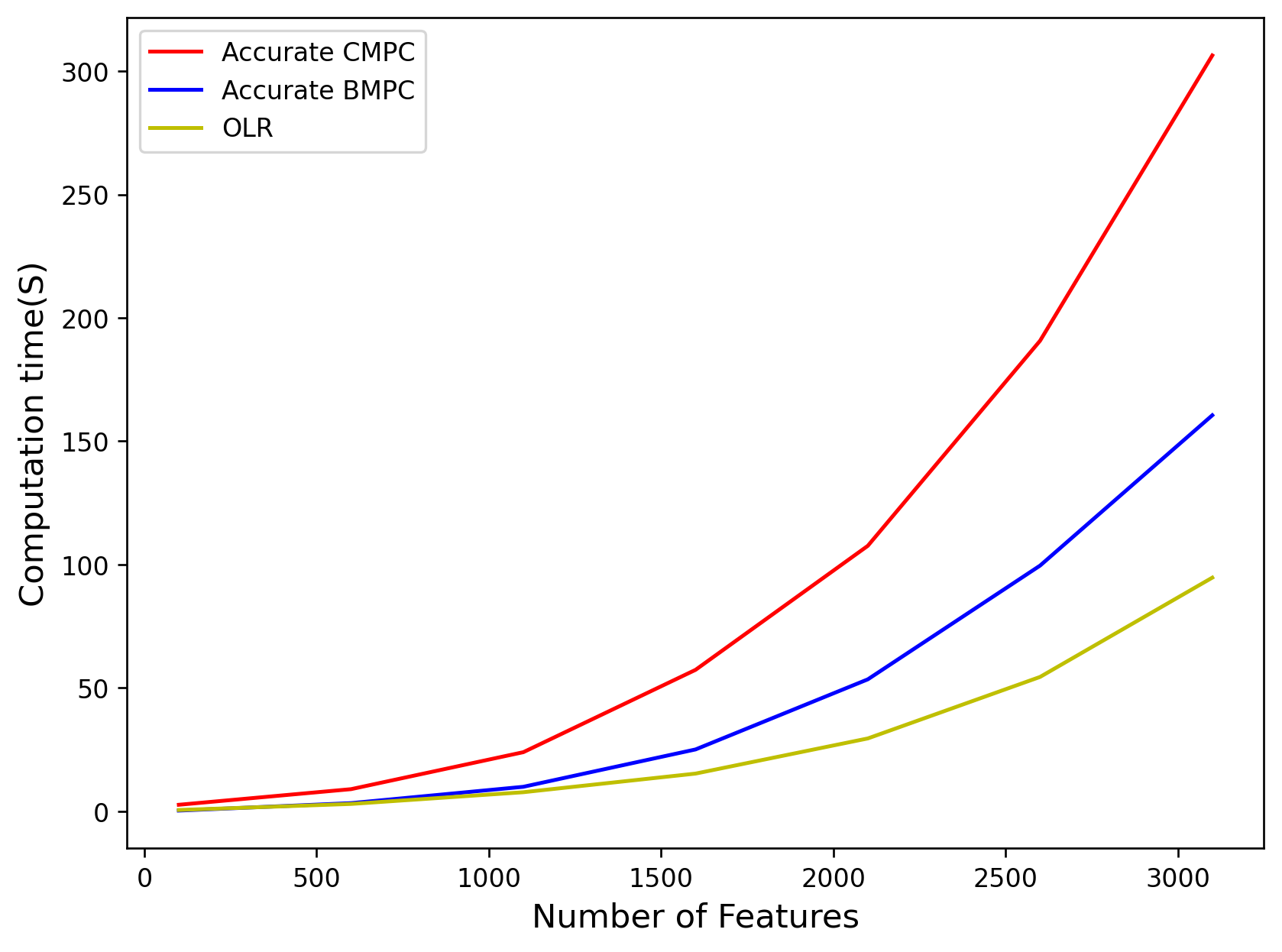}
        \caption{Efficiency comparison for increasing number of features using accurate algorithm \ref{pplr_accurate}}
        \label{fig:2}
    \end{minipage}
    \begin{minipage}[c]{0.45\textwidth}
        \centering
        \includegraphics[width=\linewidth]{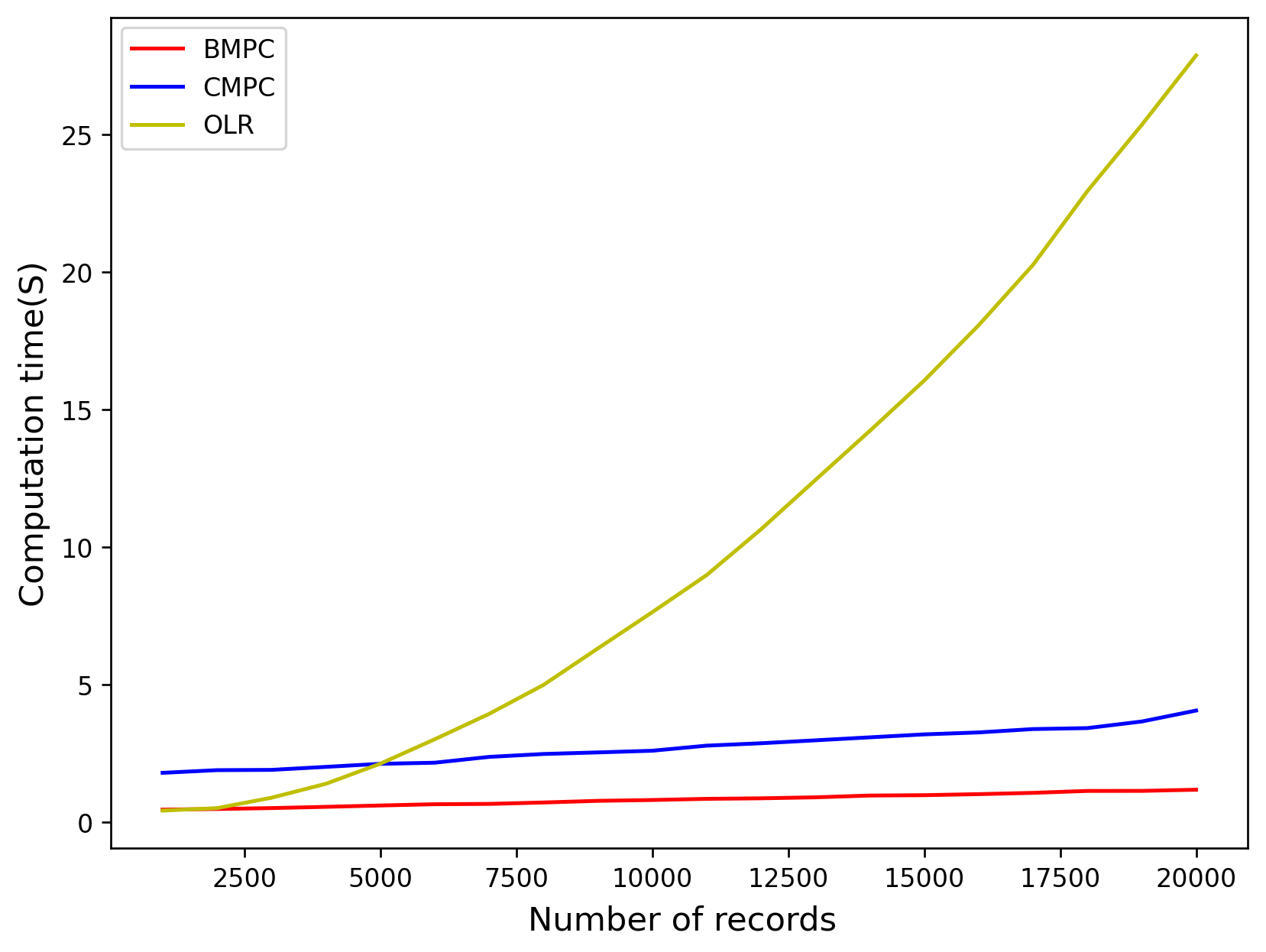}
        \caption{Efficiency comparison for increasing\\
        number of records using approximation-based\\
        algorithm \ref{pplr}}
        \label{fig:3}
    \end{minipage}
    \begin{minipage}[c]{0.45\textwidth}
        \centering
        \includegraphics[width=\linewidth]{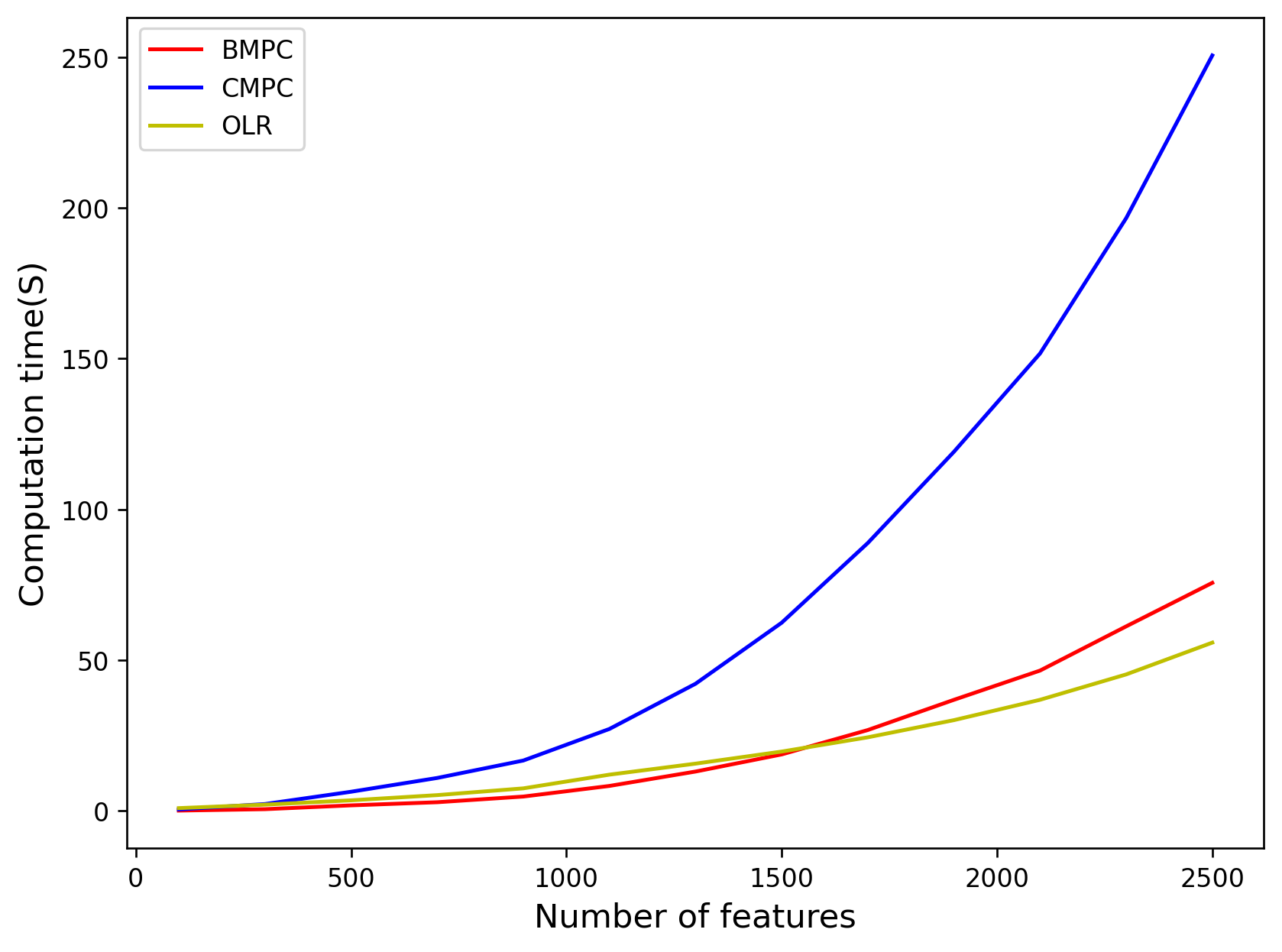}
        \caption{Efficiency comparison for increasing\\
        number of features using approximation-based\\
        algorithm \ref{pplr}}
        \label{fig:4}
    \end{minipage}
\end{figure*}

In this section, we first describe computational efficiency evaluations in terms of CPU time and memory consumption for the proposed algorithms over a real-world dataset and generated synthetic data. We finally describe the accuracy evaluations of these protocols and theoretically discuss the communication cost.

\subsection{Implementation Details}

We implemented both of the algorithms with two introduced security settings for MPC in section \ref{preliminaries} in Python. Algorithm \ref{pplr_accurate} is implemented using Beaver triple-based MPC (Accurate BMPC) and classical MPC (Accurate CMPC). The Beaver triple-based version of algorithm \ref{pplr} is called BMPC, and the classical-based MPC implementation of this algorithm is called CMPC. Moreover, to have a decent comparison, the ordinary logistic regression (OLR), which does not use MPC, is implemented.

Experiments were performed on an ARM-based M1 processor with 16GB memory, running a macOS operation system. Also, to eliminate the impact of network latency, we simulated the (distributed) computing nodes on a single computer with multiple threads. Each experiment was performed at least 10 times and reported the mean of the output. During the validation, we employed both synthetic data and real-world data sets.

We report the evaluation results concerning \textbf{computational efficiency} in terms of CPU time and memory consumption and \textbf{result accuracy}. For a fair comparison on the efficiency, we used four real-world data sets: Pima Indians Diabetes Dataset (PIMA) \cite{nr}, Low Birth Weight Study (LBW) \cite{lbw}, Prostate Cancer Study (PCS) \cite{pcs}, and Umaru Impact Study datasets (UIS)\cite{uis}. All datasets have a single binary outcome variable. Also, to satisfy the demand for large-scale studies between multiple research institutions with a large number of records, we examine our protocols with synthetic data sets of varying sizes. We generated synthetic data consists of up to 1 million records spanning up to 3000 features representing most real-world use cases. 


\begin{table*}
\centering
\caption{Memory consumption comparison of our proposed protocols using generated synthetic datasets}
\label{memory}
\begin{tabular}{|c|c|c|c|c|c|c|} 
\hline
\begin{tabular}[c]{@{}c@{}}Records\\Number\end{tabular} & \begin{tabular}[c]{@{}c@{}}Feature\\Number\end{tabular} & \begin{tabular}[c]{@{}c@{}}OLR\\\end{tabular} & BMPC & CMPC & \begin{tabular}[c]{@{}c@{}}Accurate\\BMPC~\end{tabular} & \begin{tabular}[c]{@{}c@{}}Accurate\\CMPC\end{tabular} \\ 
\hline
10000 & 50 & 3666 & 1325 & 688 & 1316 & 888 \\ 
\hline
20000 & 70 & 12559 & 1717 & 1016 & 1665 & 1325 \\ 
\hline
30000 & 90 & 22736 & 2743 & 1702 & 2654 & 2033 \\ 
\hline
40000 & 100 & 29110 & 3775 & 2337 & 3622 & 2672 \\ 
\hline
50000 & 100 & 34394 & 4575 & 2878 & 4406 & 3220 \\ 
\hline
50000 & 200 & 34512 & 8516 & 5437 & 8013 & 5770 \\
\hline
\end{tabular}
\end{table*}

\begin{table*}
\centering
\caption{A comparison between model parameters $\beta$ learned using the proposed protocols and ordinary logistic regression protocol over LBW dataset}
\label{tab_accurate_beta}
\begin{tabular}{cccccccclc} 
\toprule
\multirow{2}{*}{$\beta$} & \multirow{2}{*}{\begin{tabular}[c]{@{}c@{}}Ordinary \\LR\end{tabular}} & \multirow{2}{*}{\begin{tabular}[c]{@{}c@{}}Accurate \\BMPC\end{tabular}} & \multicolumn{3}{c}{BMPC} & \multirow{2}{*}{\begin{tabular}[c]{@{}c@{}}Accurate\\CMPC\\\end{tabular}} & \multicolumn{3}{c}{CMPC} \\ 
\cmidrule{4-6}\cline{8-10}
 &  &  & 3 & 5 & 7 &  & 3 & \multicolumn{1}{c}{5} & 7 \\ 
\hline
$\beta_1$ & 0.01574 & 0.01577 & 0.01761 & 0.01580 & 0.01480 & 0.01574 & 0.02214 & 0.01793 & 0.01630 \\
$\beta_2$ & 0.01127 & 0.01123 & 0.01171 & 0.01061 & 0.01006 & 0.01127 & 0.01534 & 0.01266 & 0.01166 \\
$\beta_3$ & 0.78666 & 0.78152 & 0.67763 & 0.62479 & 0.60392 & 0.78662 & 0.95081 & 0.81191 & 0.77010 \\
$\beta_4$ & -0.47132 & -0.46992 & -0.48975 & -0.44423 & -0.42170 & -0.47131 & -0.63960 & -0.52621 & -0.48340 \\
$\beta_5$ & -1.32410 & -1.31870 & -1.24442 & -1.13786 & -1.08974 & -1.32405 & -1.68676 & -1.41595 & -1.32408 \\
$\beta_6$ & -0.75584 & -0.75594 & -0.86971 & -0.78142 & -0.73330 & -0.75583 & -1.09894 & -0.88944 & -0.80596 \\
$\beta_7$ & -2.20748 & -2.20104 & -2.48191 & -2.23117 & -2.09511 & -2.20743 & -3.15262 & -2.56252 & -2.33208 \\
$\beta_8$ & -0.96060 & -0.95756 & -0.99906 & -0.90459 & -0.85667 & -0.96058 & -1.30317 & -1.07358 & -0.98838 \\
$\beta_9$ & -0.24569 & -0.24476 & -0.21884 & -0.20160 & -0.19465 & -0.24568 & -0.30509 & -0.25879 & -0.24367 \\
\bottomrule
\end{tabular}
\end{table*}

\subsection{Efficiency}
To compare our protocols' efficiency with an ordinary logistic regression, first, we measure the CPU time of our protocol when the number of features is constant (i.e., 250) and the number of records increases. We then calculate the CPU time of the protocols when the number of records is fixed (i.e., 7000) and the number of features increases. 

The CPU time of the proposed protocols is heavily influenced by the number of records and features of the training set. Figures \ref{fig:1} and \ref{fig:2} illustrate the CPU time of those implemented based on algorithm \ref{pplr_accurate} (Accurate BMPC and Accurate CMPC). As is shown in Figure \ref{fig:1}, Accurate BMPC has the best results when the number of records increases. This protocol computes a logistic regression model over a train set with 50000 records and 500 features in less than 15 seconds which is 20 seconds faster than Accurate CMPC protocol and 70 seconds faster than OLR. Nevertheless, increasing the number of features has a higher impact than OLR. As is shown in Figure \ref{fig:2}, by using the OLR protocol, a logistic regression model can be trained over a training set with 7000 records and 3000 features in around 90 seconds. However, computing this model using Accurate BMPC  protocol takes about 3 minutes, and using Accurate BMPC protocol requires 5 minutes.

Figures \ref{fig:3} and \ref{fig:4} illustrate the CPU time of our protocols which are implemented based on algorithm \ref{pplr} (BMP and CMPC). As shown in Figure \ref{fig:3}, both of these protocols have a better performance than OLR when the number of records increases and the number of features is fixed. Also, BMPC has a considerably better CPU time in comparison with CMPC and OLR. However, as is shown in Figure \ref{fig:4}, increasing the number of features has slightly different results. Raising the number of features decreases the efficiency of all three protocols. CMPC receives the highest impact from rising the number of features, but BMPC still has an acceptable efficiency level. For example, BMPC can train a model with 7000 records and 2500 features in less than one minute, which is 7 seconds higher than OLR, and four times better than CMPC. Therefore, we can conclude that CMPC is not the right choice when we have a dataset with a considerable number of features.

Besides, to measure the efficiency in terms of memory consumption, a python module called ``memory\_profiler'' \cite{memory} has been utilized. Table \ref{memory} indicates memory consumption for different introduced protocols and ordinary logistic regression. All implemented protocols in both security settings consume remarkably less memory during the training process. As is describes in the table \ref{memory}, to train a logistic regression model over a dataset with 50000 records and 200 features, the OLR protocol requires about 30 GB of memory. However, the BMPC and Accurate BMPC about 8 GB of memory to train this model. Also, the CMPC and Accurate CMPC protocols have a better consumption rate than BMPC, and less than 5 GB of memory is needed to train such a model.

\begin{table*}
\centering
\caption{Accuracy comparison result for real-word datasets with different settings}
\label{accuracy_table}
\begin{tabular}{|c|c|c|c|c|c|c|c|c|c|} 
\cline{5-10}
\multicolumn{1}{c}{} & \multicolumn{1}{l}{} & \multicolumn{1}{l}{} &  & \multicolumn{2}{c|}{\begin{tabular}[c]{@{}c@{}}CMPC \end{tabular}} & \multicolumn{2}{c|}{\begin{tabular}[c]{@{}c@{}}BMPC \end{tabular}} & \multicolumn{2}{c|}{OLR} \\ 
\hline
Dataset & \begin{tabular}[c]{@{}c@{}}Records\\Num \end{tabular} & \begin{tabular}[c]{@{}c@{}}Feature\\Num \end{tabular} & \begin{tabular}[c]{@{}c@{}}g(x) \\degree \end{tabular} & Accuracy & AUC & \multicolumn{1}{l|}{Accuracy} & AUC & \multicolumn{1}{r|}{Accuracy} & \multicolumn{1}{l|}{AUC} \\ 
\hline
 & \multicolumn{1}{l|}{} & \multicolumn{1}{l|}{} & 3 & 71.87\% & \multicolumn{1}{r|}{0.740} & 71.87\% & 0.740 & 71.87\% & 0.740 \\ 
\cline{4-10}
\multirow{2}{*}{PIMA} & \multirow{2}{*}{768} & \multirow{2}{*}{9} & 5 & 71.87\% & \multicolumn{1}{r|}{0.741} & 71.87\% & 0.740 & 71.87\% & 0.740 \\ 
\cline{4-10}
 &  &  & 7 & 71.87\% & \multicolumn{1}{l|}{0.741} & 71.87\% & 0.741 & 71.87\% & 0.741 \\ 
\cline{4-10}
\multicolumn{1}{|l|}{} & \multicolumn{1}{l|}{} & \multicolumn{1}{l|}{} & No approx & - & - & - & - & 71.87\% & 0.741 \\ 
\hline
 & \multicolumn{1}{l|}{} & \multicolumn{1}{l|}{} & 3 & 81.05\% & \multicolumn{1}{r|}{0.842} & 81.05\% & 0.846 & 80\% & 0.846 \\ 
\cline{4-10}
\multirow{2}{*}{PCS} & \multirow{2}{*}{379} & \multirow{2}{*}{10} & 5 & 81.05\% & \multicolumn{1}{r|}{0.845} & 81.05\% & 0.847 & 80\% & 0.847 \\ 
\cline{4-10}
 &  &  & 7 & 81.05\% & \multicolumn{1}{l|}{0.847} & 81.05\% & 0.848 & 81.05\% & 0.848 \\ 
\cline{4-10}
 & \multicolumn{1}{l|}{} & \multicolumn{1}{l|}{} & No approx & - & - & - & - & 81.05\% & 0.848 \\ 
\hline
 & \multicolumn{1}{l|}{} & \multicolumn{1}{l|}{} & 3 & 64.58\% & 0.519 & 64.58\% & 0.519 & 64.58\% & \multicolumn{1}{l|}{0.519} \\ 
\cline{4-10}
\multirow{2}{*}{LBW} & \multirow{2}{*}{189} & \multirow{2}{*}{10} & 5 & 64.58\% & \multicolumn{1}{r|}{0.519} & 64.58\% & 0.519 & 64.58\% & \multicolumn{1}{l|}{0.519} \\ 
\cline{4-10}
 &  &  & 7 & 62.5\% & \multicolumn{1}{l|}{0.519} & 62.5\% & 0.517 & 64.58\% & \multicolumn{1}{l|}{0.519} \\ 
\cline{4-10}
 & \multicolumn{1}{l|}{} & \multicolumn{1}{l|}{} & No approx & - & - & - & - & 62.5\% & 0.523 \\ 
\hline
 & \multicolumn{1}{l|}{} & \multicolumn{1}{l|}{} & 3 & 73.61\% & 0.651 & 73.61\% & 0.651 & 73.61\% & \multicolumn{1}{l|}{0.651} \\ 
\cline{4-10}
\multirow{2}{*}{UIS} & \multirow{2}{*}{575} & \multirow{2}{*}{9} & 5 & 72.91\% & \multicolumn{1}{r|}{0.652} & 72.91\% & 0.652 & 72.91\% & \multicolumn{1}{l|}{0.652} \\ 
\cline{4-10}
 &  &  & 7 & 72.91\% & 0.655 & 73.61\% & \multicolumn{1}{l|}{0.651} & 72.91\% & \multicolumn{1}{l|}{0.655} \\ 
\cline{4-10}
 & \multicolumn{1}{l|}{} & \multicolumn{1}{l|}{} & No approx & - & - & - & - & 72.22\% & 0.656 \\
\hline
\end{tabular}
\end{table*}

\subsection{Accuracy}

One of the main subjects that we considered in examining our protocols is whether they are accurate. We measured the accuracy based on the estimated model parameters' precision during the training phase over the Low Birth Weight Study dataset. To do this, we compared the obtained vector of coefficient $\beta$ from our protocols with the ones estimated using the OLR protocols. As Table \ref{tab_accurate_beta} presents, the Accurate BMPC and Accurate CMPC protocols' model parameters are almost the same as the model parameters estimated using OLR protocols. Moreover, the model parameters estimated from protocols based on the algorithm \ref{pplr} (BMPC and CMPC), which employ various approximations, have an acceptable level of accuracy compared to the model parameters estimated using OLR protocols.

Since we introduced several approximation schemes in BMPC and CMPC protocols (e.g., fixed-point Hessian matrix, the least-squares approximation of the Sigmoid function, and matrix inversion), and to have a better understanding of the accuracy level of these protocols, we compare the prediction accuracy achieved by these protocols with that obtained from OLR. To do this, we calculated the percentage (\%) of the correct predictions of the models produced on four different datasets (25\% of training samples were assigned to the test set) in different settings based on the degree of Sigmoid function approximation.  All the accuracy measurement results are summarized in Table \ref{accuracy_table}. This table presents the average prediction accuracy percentage when threshold = 0.5 and the AUC (Area Under the Receiver Operating Characteristic Curve), which estimates a binary classifier's quality. It is clear from the table data that the approximations used in BMPC and CMPC protocols do not significantly affect the estimated model's accuracy. In other words, although based on the information provided in Table \ref{tab_accurate_beta}, varying the Sigmoid approximation degrees affect the exactness of model parameters learned during the training phase, these differences do not considerably impact the classification accuracy over the chosen datasets.

\subsection{Communication cost}

The efficiency of multi-party computation protocols in terms of communication costs is considered a fundamental concern. Since in our implementation, we simulate each computation party as a thread in a multi-threading setting, we theoretically compute the communication cost in this part. To do this, first, we compute the primary operations' communication cost, such as addition and multiplication, in both honest majority and dishonest majority settings. Secondly, we compute the total communication cost for the whole protocol based on the number of primary operations contained.

The addition operation in both security settings does not require any communication between the parties. However, multiplication requires multiple communication rounds. In the honest majority setting and with three computation parties, Bogdanov \cite{bogdanov2013sharemind} explained that each time performing the MPC-based multiplication requires exchanging 15 messages between the computation parties. If we consider each message with the size of 32 bits, the communication cost for one-time multiplication will be 420 bits. Besides, one time executing our logistic regression protocol (CMPC) requires performing between 100 to 300 times multiplication protocol (based on the degree of least-squares Sigmoid approximation and choosing a good start value for the inversion operation). Therefore, to compute the logistic regression model in an honest majority setting with our protocol, at least 42 Kb data will be exchanged.

In the dishonest majority setting and using the Beaver multiplication approach, communication costs are lower than the honest majority setting. The multiplication procedure in this method is split into the offline and online phases. During the offline phase, multiplication triples will be generated and distributed before the computation parties' inputs be associated. Therefore, the communication cost of this phase can be safely ignored. During the online phase and in the two-party setting, each computation party sends only two messages to the other party to perform the multiplication. Accordingly, if we consider each message with the size of 32 bits, therefore 128-bits data require to be exchanged for performing one-time multiplication in this setting, which is 3.75 times less than the honest majority setting. For the whole logistic regression protocol, 12.5 Kb of data will be transferred.

%% file: conclusions.tex
\section{Conclusions}

There is an increasing interest in applying machine learning algorithms to sensitive data, such as medical data. In this paper, we described novel protocols for implementing secure and private logistic regression training among distributed parties using multi-party computation protocols. We evaluated the performance of our protocols through experiments on real-world and synthetic datasets. With the latter, we showed that our solutions scale well when apply to a dataset with a very large number of records and features. Our experiments also showed that our protocol achieves high accuracy while maintaining a reasonable level of efficiency. In the future, we will extend our protocols to support secure and efficient multi-class logistic regression.




%% file: main.bbl

\begin{thebibliography}{33}
\ifx \bisbn   \undefined \def \bisbn  #1{ISBN #1}\fi
\ifx \binits  \undefined \def \binits#1{#1}\fi
\ifx \bauthor  \undefined \def \bauthor#1{#1}\fi
\ifx \batitle  \undefined \def \batitle#1{#1}\fi
\ifx \bjtitle  \undefined \def \bjtitle#1{#1}\fi
\ifx \bvolume  \undefined \def \bvolume#1{\textbf{#1}}\fi
\ifx \byear  \undefined \def \byear#1{#1}\fi
\ifx \bissue  \undefined \def \bissue#1{#1}\fi
\ifx \bfpage  \undefined \def \bfpage#1{#1}\fi
\ifx \blpage  \undefined \def \blpage #1{#1}\fi
\ifx \burl  \undefined \def \burl#1{\textsf{#1}}\fi
\ifx \doiurl  \undefined \def \doiurl#1{\textsf{#1}}\fi
\ifx \betal  \undefined \def \betal{\textit{et al.}}\fi
\ifx \binstitute  \undefined \def \binstitute#1{#1}\fi
\ifx \binstitutionaled  \undefined \def \binstitutionaled#1{#1}\fi
\ifx \bctitle  \undefined \def \bctitle#1{#1}\fi
\ifx \beditor  \undefined \def \beditor#1{#1}\fi
\ifx \bpublisher  \undefined \def \bpublisher#1{#1}\fi
\ifx \bbtitle  \undefined \def \bbtitle#1{#1}\fi
\ifx \bedition  \undefined \def \bedition#1{#1}\fi
\ifx \bseriesno  \undefined \def \bseriesno#1{#1}\fi
\ifx \blocation  \undefined \def \blocation#1{#1}\fi
\ifx \bsertitle  \undefined \def \bsertitle#1{#1}\fi
\ifx \bsnm \undefined \def \bsnm#1{#1}\fi
\ifx \bsuffix \undefined \def \bsuffix#1{#1}\fi
\ifx \bparticle \undefined \def \bparticle#1{#1}\fi
\ifx \barticle \undefined \def \barticle#1{#1}\fi
\ifx \bconfdate \undefined \def \bconfdate #1{#1}\fi
\ifx \botherref \undefined \def \botherref #1{#1}\fi
\ifx \url \undefined \def \url#1{\textsf{#1}}\fi
\ifx \bchapter \undefined \def \bchapter#1{#1}\fi
\ifx \bbook \undefined \def \bbook#1{#1}\fi
\ifx \bcomment \undefined \def \bcomment#1{#1}\fi
\ifx \oauthor \undefined \def \oauthor#1{#1}\fi
\ifx \citeauthoryear \undefined \def \citeauthoryear#1{#1}\fi
\ifx \endbibitem  \undefined \def \endbibitem {}\fi
\ifx \bconflocation  \undefined \def \bconflocation#1{#1}\fi
\ifx \arxivurl  \undefined \def \arxivurl#1{\textsf{#1}}\fi
\csname PreBibitemsHook\endcsname

\bibitem{hosmer2013applied}
\begin{bchapter}
\bauthor{\bsnm{Hosmer~Jr}, \binits{D.W.}},
\bauthor{\bsnm{Lemeshow}, \binits{S.}},
\bauthor{\bsnm{Sturdivant}, \binits{R.X.}}:
\bctitle{Applied logistic regression}.
(\byear{2013}).
\bcomment{John Wiley \& Sons}
\end{bchapter}
\endbibitem

\bibitem{boxwala2011using}
\begin{barticle}
\bauthor{\bsnm{Boxwala}, \binits{A.A.}},
\bauthor{\bsnm{Kim}, \binits{J.}},
\bauthor{\bsnm{Grillo}, \binits{J.M.}},
\bauthor{\bsnm{Ohno-Machado}, \binits{L.}}:
\batitle{Using statistical and machine learning to help institutions detect
  suspicious access to electronic health records}.
\bjtitle{Journal of the American Medical Informatics Association}
\bvolume{18}(\bissue{4}),
\bfpage{498}--\blpage{505}
(\byear{2011})
\end{barticle}
\endbibitem

\bibitem{riley2020calculating}
\begin{botherref}
\oauthor{\bsnm{Riley}, \binits{R.D.}},
\oauthor{\bsnm{Ensor}, \binits{J.}},
\oauthor{\bsnm{Snell}, \binits{K.I.}},
\oauthor{\bsnm{Harrell}, \binits{F.E.}},
\oauthor{\bsnm{Martin}, \binits{G.P.}},
\oauthor{\bsnm{Reitsma}, \binits{J.B.}},
\oauthor{\bsnm{Moons}, \binits{K.G.}},
\oauthor{\bsnm{Collins}, \binits{G.}},
\oauthor{\bparticle{van} \bsnm{Smeden}, \binits{M.}}:
Calculating the sample size required for developing a clinical prediction
  model.
Bmj
\textbf{368}
(2020)
\end{botherref}
\endbibitem

\bibitem{jagadeesh2017deriving}
\begin{barticle}
\bauthor{\bsnm{Jagadeesh}, \binits{K.A.}},
\bauthor{\bsnm{Wu}, \binits{D.J.}},
\bauthor{\bsnm{Birgmeier}, \binits{J.A.}},
\bauthor{\bsnm{Boneh}, \binits{D.}},
\bauthor{\bsnm{Bejerano}, \binits{G.}}:
\batitle{Deriving genomic diagnoses without revealing patient genomes}.
\bjtitle{Science}
\bvolume{357}(\bissue{6352}),
\bfpage{692}--\blpage{695}
(\byear{2017})
\end{barticle}
\endbibitem

\bibitem{shi2016secure}
\begin{barticle}
\bauthor{\bsnm{Shi}, \binits{H.}},
\bauthor{\bsnm{Jiang}, \binits{C.}},
\bauthor{\bsnm{Dai}, \binits{W.}},
\bauthor{\bsnm{Jiang}, \binits{X.}},
\bauthor{\bsnm{Tang}, \binits{Y.}},
\bauthor{\bsnm{Ohno-Machado}, \binits{L.}},
\bauthor{\bsnm{Wang}, \binits{S.}}:
\batitle{Secure multi-party computation grid logistic regression (smac-glore)}.
\bjtitle{BMC medical informatics and decision making}
\bvolume{16}(\bissue{3}),
\bfpage{89}
(\byear{2016})
\end{barticle}
\endbibitem

\bibitem{xie2016privlogit}
\begin{botherref}
\oauthor{\bsnm{Xie}, \binits{W.}},
\oauthor{\bsnm{Wang}, \binits{Y.}},
\oauthor{\bsnm{Boker}, \binits{S.M.}},
\oauthor{\bsnm{Brown}, \binits{D.E.}}:
Privlogit: Efficient privacy-preserving logistic regression by tailoring
  numerical optimizers.
arXiv preprint arXiv:1611.01170
(2016)
\end{botherref}
\endbibitem

\bibitem{pop00001}
\begin{barticle}
\bauthor{\bsnm{De~Cock}, \binits{M.}},
\bauthor{\bsnm{Dowsley}, \binits{R.}},
\bauthor{\bsnm{Horst}, \binits{C.}},
\bauthor{\bsnm{Katti}, \binits{R.}},
\bauthor{\bsnm{Nascimento}, \binits{A.C.}},
\bauthor{\bsnm{Poon}, \binits{W.-S.}},
\bauthor{\bsnm{Truex}, \binits{S.}}:
\batitle{Efficient and private scoring of decision trees, support vector
  machines and logistic regression models based on pre-computation}.
\bjtitle{IEEE Transactions on Dependable and Secure Computing}
\bvolume{16}(\bissue{2}),
\bfpage{217}--\blpage{230}
(\byear{2017})
\end{barticle}
\endbibitem

\bibitem{beaver1997commodity}
\begin{bchapter}
\bauthor{\bsnm{Beaver}, \binits{D.}}:
\bctitle{Commodity-based cryptography}.
In: \bbtitle{Proceedings of the Twenty-ninth Annual ACM Symposium on Theory of
  Computing},
pp. \bfpage{446}--\blpage{455}
(\byear{1997})
\end{bchapter}
\endbibitem

\bibitem{mohassel2017secureml}
\begin{bchapter}
\bauthor{\bsnm{Mohassel}, \binits{P.}},
\bauthor{\bsnm{Zhang}, \binits{Y.}}:
\bctitle{Secureml: A system for scalable privacy-preserving machine learning}.
In: \bbtitle{2017 IEEE Symposium on Security and Privacy (SP)},
pp. \bfpage{19}--\blpage{38}
(\byear{2017}).
\bcomment{IEEE}
\end{bchapter}
\endbibitem

\bibitem{gentry2009fully}
\begin{botherref}
\oauthor{\bsnm{Gentry}, \binits{C.}},
\oauthor{\bsnm{Boneh}, \binits{D.}}:
A Fully Homomorphic Encryption Scheme
vol. 20.
Stanford university Stanford
\end{botherref}
\endbibitem

\bibitem{pop00026}
\begin{bchapter}
\bauthor{\bsnm{Yoo}, \binits{J.S.}},
\bauthor{\bsnm{Hwang}, \binits{J.H.}},
\bauthor{\bsnm{Song}, \binits{B.K.}},
\bauthor{\bsnm{Yoon}, \binits{J.W.}}:
\bctitle{A bitwise logistic regression using binary approximation and real
  number division in homomorphic encryption scheme}.
In: \bbtitle{International Conference on Information Security Practice and
  Experience},
pp. \bfpage{20}--\blpage{40}
(\byear{2019}).
\bcomment{Springer}
\end{bchapter}
\endbibitem

\bibitem{pop00007}
\begin{botherref}
\oauthor{\bsnm{MLD}, \binits{R.}},
\oauthor{\bsnm{Fienberg}, \binits{S.}},
\oauthor{\bsnm{Nardi}, \binits{Y.}}:
Secure multiparty linear and logistic regression based on homomorphic
  encryption.
cs.cmu.edu.
Query date: 2020-06-24 08:59:23
\end{botherref}
\endbibitem

\bibitem{pop00015}
\begin{barticle}
\bauthor{\bsnm{Carpov}, \binits{S.}},
\bauthor{\bsnm{Gama}, \binits{N.}},
\bauthor{\bsnm{Georgieva}, \binits{M.}},
\bauthor{\bsnm{Troncoso-Pastoriza}, \binits{J.R.}}:
\batitle{Privacy-preserving semi-parallel logistic regression training with
  fully homomorphic encryption}.
\bjtitle{BMC Medical Genomics}
\bvolume{13}(\bissue{7}),
\bfpage{1}--\blpage{10}
(\byear{2020})
\end{barticle}
\endbibitem

\bibitem{pop00011}
\begin{barticle}
\bauthor{\bsnm{Kim}, \binits{M.}},
\bauthor{\bsnm{Song}, \binits{Y.}},
\bauthor{\bsnm{Wang}, \binits{S.}},
\bauthor{\bsnm{Xia}, \binits{Y.}},
\bauthor{\bsnm{Jiang}, \binits{X.}}:
\batitle{Secure logistic regression based on homomorphic encryption: Design and
  evaluation}.
\bjtitle{JMIR medical informatics}
\bvolume{6}(\bissue{2}),
\bfpage{19}
(\byear{2018})
\end{barticle}
\endbibitem

\bibitem{pop00020}
\begin{bchapter}
\bauthor{\bsnm{Han}, \binits{K.}},
\bauthor{\bsnm{Hong}, \binits{S.}},
\bauthor{\bsnm{Cheon}, \binits{J.H.}},
\bauthor{\bsnm{Park}, \binits{D.}}:
\bctitle{Logistic regression on homomorphic encrypted data at scale}.
In: \bbtitle{Proceedings of the AAAI Conference on Artificial Intelligence},
vol. \bseriesno{33},
pp. \bfpage{9466}--\blpage{9471}
(\byear{2019})
\end{bchapter}
\endbibitem

\bibitem{pop00021}
\begin{botherref}
\oauthor{\bsnm{Han}, \binits{K.}},
\oauthor{\bsnm{Hong}, \binits{S.}},
\oauthor{\bsnm{Cheon}, \binits{J.}},
\oauthor{\bsnm{Park}, \binits{D.}}:
Efficient logistic regression on large encrypted data.
IACR Cryptol. ePrint Arch.
(2018).
Query date: 2020-06-24 08:59:23
\end{botherref}
\endbibitem

\bibitem{pop00008}
\begin{bbook}
\bauthor{\bsnm{Djonatan}, \binits{P.}}:
\bbtitle{Privacy-preserving Analytics: Secure Logistic Regression},
(\byear{2019}).
\bcomment{Query date: 2020-06-24 08:59:23}.
\burl{https://dr.ntu.edu.sg/handle/10356/77126}
\end{bbook}
\endbibitem

\bibitem{pop00019}
\begin{bchapter}
\bauthor{\bsnm{Du}, \binits{W.}},
\bauthor{\bsnm{Li}, \binits{A.}},
\bauthor{\bsnm{Li}, \binits{Q.}}:
\bctitle{Privacy-preserving multiparty learning for logistic regression}.
In: \bbtitle{International Conference on Security and Privacy in Communication
  Systems},
pp. \bfpage{549}--\blpage{568}
(\byear{2018}).
\bcomment{Springer}
\end{bchapter}
\endbibitem

\bibitem{chaudhuri2009privacy}
\begin{bchapter}
\bauthor{\bsnm{Chaudhuri}, \binits{K.}},
\bauthor{\bsnm{Monteleoni}, \binits{C.}}:
\bctitle{Privacy-preserving logistic regression}.
In: \bbtitle{Advances in Neural Information Processing Systems},
pp. \bfpage{289}--\blpage{296}
(\byear{2009})
\end{bchapter}
\endbibitem

\bibitem{el2013secure}
\begin{barticle}
\bauthor{\bsnm{El~Emam}, \binits{K.}},
\bauthor{\bsnm{Samet}, \binits{S.}},
\bauthor{\bsnm{Arbuckle}, \binits{L.}},
\bauthor{\bsnm{Tamblyn}, \binits{R.}},
\bauthor{\bsnm{Earle}, \binits{C.}},
\bauthor{\bsnm{Kantarcioglu}, \binits{M.}}:
\batitle{A secure distributed logistic regression protocol for the detection of
  rare adverse drug events}.
\bjtitle{Journal of the American Medical Informatics Association}
\bvolume{20}(\bissue{3}),
\bfpage{453}--\blpage{461}
(\byear{2013})
\end{barticle}
\endbibitem

\bibitem{kim2019secure}
\begin{barticle}
\bauthor{\bsnm{Kim}, \binits{M.}},
\bauthor{\bsnm{Lee}, \binits{J.}},
\bauthor{\bsnm{Ohno-Machado}, \binits{L.}},
\bauthor{\bsnm{Jiang}, \binits{X.}}:
\batitle{Secure and differentially private logistic regression for horizontally
  distributed data}.
\bjtitle{IEEE Transactions on Information Forensics and Security}
\bvolume{15},
\bfpage{695}--\blpage{710}
(\byear{2019})
\end{barticle}
\endbibitem

\bibitem{bogdanov2012high}
\begin{barticle}
\bauthor{\bsnm{Bogdanov}, \binits{D.}},
\bauthor{\bsnm{Niitsoo}, \binits{M.}},
\bauthor{\bsnm{Toft}, \binits{T.}},
\bauthor{\bsnm{Willemson}, \binits{J.}}:
\batitle{High-performance secure multi-party computation for data mining
  applications}.
\bjtitle{International Journal of Information Security}
\bvolume{11}(\bissue{6}),
\bfpage{403}--\blpage{418}
(\byear{2012})
\end{barticle}
\endbibitem

\bibitem{beaver1991efficient}
\begin{bchapter}
\bauthor{\bsnm{Beaver}, \binits{D.}}:
\bctitle{Efficient multiparty protocols using circuit randomization}.
In: \bbtitle{Annual International Cryptology Conference},
pp. \bfpage{420}--\blpage{432}
(\byear{1991}).
\bcomment{Springer}
\end{bchapter}
\endbibitem

\bibitem{nardi2012achieving}
\begin{botherref}
\oauthor{\bsnm{Nardi}, \binits{Y.}},
\oauthor{\bsnm{Fienberg}, \binits{S.E.}},
\oauthor{\bsnm{Hall}, \binits{R.J.}}:
Achieving both valid and secure logistic regression analysis on aggregated data
  from different private sources.
Journal of Privacy and Confidentiality
\textbf{4}(1)
(2012)
\end{botherref}
\endbibitem

\bibitem{agresti2003categorical}
\begin{botherref}
\oauthor{\bsnm{Agresti}, \binits{A.}}:
Categorical data analysis
\textbf{482}
(2003)
\end{botherref}
\endbibitem

\bibitem{kim2018secure}
\begin{barticle}
\bauthor{\bsnm{Kim}, \binits{M.}},
\bauthor{\bsnm{Song}, \binits{Y.}},
\bauthor{\bsnm{Wang}, \binits{S.}},
\bauthor{\bsnm{Xia}, \binits{Y.}},
\bauthor{\bsnm{Jiang}, \binits{X.}}:
\batitle{Secure logistic regression based on homomorphic encryption: Design and
  evaluation}.
\bjtitle{JMIR medical informatics}
\bvolume{6}(\bissue{2}),
\bfpage{19}
(\byear{2018})
\end{barticle}
\endbibitem

\bibitem{bohning1999lower}
\begin{barticle}
\bauthor{\bsnm{B{\"o}hning}, \binits{D.}}:
\batitle{The lower bound method in probit regression}.
\bjtitle{Computational statistics \& data analysis}
\bvolume{30}(\bissue{1}),
\bfpage{13}--\blpage{17}
(\byear{1999})
\end{barticle}
\endbibitem

\bibitem{nr}
\begin{botherref}
\oauthor{\bsnm{Dua}, \binits{D.}},
\oauthor{\bsnm{Graff}, \binits{C.}}:
{UCI} Machine Learning Repository
(2017).
\url{http://archive.ics.uci.edu/ml}
\end{botherref}
\endbibitem

\bibitem{lbw}
\begin{botherref}
lbw: Low Birth Weight study data
(2019).
\url{https://rdrr.io/rforge/LogisticDx/man/lbw.html}
\end{botherref}
\endbibitem

\bibitem{pcs}
\begin{botherref}
pcs: Prostate Cancer Study data
(2019).
\url{https://rdrr.io/rforge/LogisticDx/man/pcs.html}
\end{botherref}
\endbibitem

\bibitem{uis}
\begin{botherref}
uis: UMARU IMPACT Study data
(2019).
\url{https://rdrr.io/rforge/LogisticDx/man/uis.html}
\end{botherref}
\endbibitem

\bibitem{memory}
\begin{botherref}
memory-profiler
(2021).
\url{https://pypi.org/project/memory-profiler/}
\end{botherref}
\endbibitem

\bibitem{bogdanov2013sharemind}
\begin{botherref}
\oauthor{\bsnm{Bogdanov}, \binits{D.}}:
Sharemind: programmable secure computations with practical applications.
PhD thesis,
Tartu University
(2013)
\end{botherref}
\endbibitem

\end{thebibliography}

\newcommand{\BMCxmlcomment}[1]{}

\BMCxmlcomment{

<refgrp>

<bibl id="B1">
  <title><p>Applied logistic regression</p></title>
  <aug>
    <au><snm>Hosmer Jr</snm><fnm>DW</fnm></au>
    <au><snm>Lemeshow</snm><fnm>S</fnm></au>
    <au><snm>Sturdivant</snm><fnm>RX</fnm></au>
  </aug>
  <pubdate>2013</pubdate>
</bibl>

<bibl id="B2">
  <title><p>Using statistical and machine learning to help institutions detect
  suspicious access to electronic health records</p></title>
  <aug>
    <au><snm>Boxwala</snm><fnm>AA</fnm></au>
    <au><snm>Kim</snm><fnm>J</fnm></au>
    <au><snm>Grillo</snm><fnm>JM</fnm></au>
    <au><snm>Ohno Machado</snm><fnm>L</fnm></au>
  </aug>
  <source>Journal of the American Medical Informatics Association</source>
  <publisher>BMJ Group</publisher>
  <pubdate>2011</pubdate>
  <volume>18</volume>
  <issue>4</issue>
  <fpage>498</fpage>
  <lpage>-505</lpage>
</bibl>

<bibl id="B3">
  <title><p>Calculating the sample size required for developing a clinical
  prediction model</p></title>
  <aug>
    <au><snm>Riley</snm><fnm>RD</fnm></au>
    <au><snm>Ensor</snm><fnm>J</fnm></au>
    <au><snm>Snell</snm><fnm>KI</fnm></au>
    <au><snm>Harrell</snm><fnm>FE</fnm></au>
    <au><snm>Martin</snm><fnm>GP</fnm></au>
    <au><snm>Reitsma</snm><fnm>JB</fnm></au>
    <au><snm>Moons</snm><fnm>KG</fnm></au>
    <au><snm>Collins</snm><fnm>G</fnm></au>
    <au><snm>Smeden</snm><fnm>M</fnm></au>
  </aug>
  <source>Bmj</source>
  <publisher>British Medical Journal Publishing Group</publisher>
  <pubdate>2020</pubdate>
  <volume>368</volume>
</bibl>

<bibl id="B4">
  <title><p>Deriving genomic diagnoses without revealing patient
  genomes</p></title>
  <aug>
    <au><snm>Jagadeesh</snm><fnm>KA</fnm></au>
    <au><snm>Wu</snm><fnm>DJ</fnm></au>
    <au><snm>Birgmeier</snm><fnm>JA</fnm></au>
    <au><snm>Boneh</snm><fnm>D</fnm></au>
    <au><snm>Bejerano</snm><fnm>G</fnm></au>
  </aug>
  <source>Science</source>
  <publisher>American Association for the Advancement of Science</publisher>
  <pubdate>2017</pubdate>
  <volume>357</volume>
  <issue>6352</issue>
  <fpage>692</fpage>
  <lpage>-695</lpage>
</bibl>

<bibl id="B5">
  <title><p>Secure multi-pArty computation grid LOgistic REgression
  (SMAC-GLORE)</p></title>
  <aug>
    <au><snm>Shi</snm><fnm>H</fnm></au>
    <au><snm>Jiang</snm><fnm>C</fnm></au>
    <au><snm>Dai</snm><fnm>W</fnm></au>
    <au><snm>Jiang</snm><fnm>X</fnm></au>
    <au><snm>Tang</snm><fnm>Y</fnm></au>
    <au><snm>Ohno Machado</snm><fnm>L</fnm></au>
    <au><snm>Wang</snm><fnm>S</fnm></au>
  </aug>
  <source>BMC medical informatics and decision making</source>
  <publisher>Springer</publisher>
  <pubdate>2016</pubdate>
  <volume>16</volume>
  <issue>3</issue>
  <fpage>89</fpage>
</bibl>

<bibl id="B6">
  <title><p>Privlogit: Efficient privacy-preserving logistic regression by
  tailoring numerical optimizers</p></title>
  <aug>
    <au><snm>Xie</snm><fnm>W</fnm></au>
    <au><snm>Wang</snm><fnm>Y</fnm></au>
    <au><snm>Boker</snm><fnm>SM</fnm></au>
    <au><snm>Brown</snm><fnm>DE</fnm></au>
  </aug>
  <source>arXiv preprint arXiv:1611.01170</source>
  <pubdate>2016</pubdate>
</bibl>

<bibl id="B7">
  <title><p>Efficient and private scoring of decision trees, support vector
  machines and logistic regression models based on pre-computation</p></title>
  <aug>
    <au><snm>De Cock</snm><fnm>M</fnm></au>
    <au><snm>Dowsley</snm><fnm>R</fnm></au>
    <au><snm>Horst</snm><fnm>C</fnm></au>
    <au><snm>Katti</snm><fnm>R</fnm></au>
    <au><snm>Nascimento</snm><fnm>AC</fnm></au>
    <au><snm>Poon</snm><fnm>WS</fnm></au>
    <au><snm>Truex</snm><fnm>S</fnm></au>
  </aug>
  <source>IEEE Transactions on Dependable and Secure Computing</source>
  <publisher>IEEE</publisher>
  <pubdate>2017</pubdate>
  <volume>16</volume>
  <issue>2</issue>
  <fpage>217</fpage>
  <lpage>-230</lpage>
</bibl>

<bibl id="B8">
  <title><p>Commodity-based cryptography</p></title>
  <aug>
    <au><snm>Beaver</snm><fnm>D</fnm></au>
  </aug>
  <source>Proceedings of the twenty-ninth annual ACM symposium on Theory of
  computing</source>
  <pubdate>1997</pubdate>
  <fpage>446</fpage>
  <lpage>-455</lpage>
</bibl>

<bibl id="B9">
  <title><p>Secureml: A system for scalable privacy-preserving machine
  learning</p></title>
  <aug>
    <au><snm>Mohassel</snm><fnm>P</fnm></au>
    <au><snm>Zhang</snm><fnm>Y</fnm></au>
  </aug>
  <source>2017 IEEE Symposium on Security and Privacy (SP)</source>
  <pubdate>2017</pubdate>
  <fpage>19</fpage>
  <lpage>-38</lpage>
</bibl>

<bibl id="B10">
  <title><p>A fully homomorphic encryption scheme</p></title>
  <aug>
    <au><snm>Gentry</snm><fnm>C</fnm></au>
    <au><snm>Boneh</snm><fnm>D</fnm></au>
  </aug>
  <publisher>Stanford university Stanford</publisher>
  <volume>20</volume>
  <issue>9</issue>
</bibl>

<bibl id="B11">
  <title><p>A bitwise logistic regression using binary approximation and real
  number division in homomorphic encryption scheme</p></title>
  <aug>
    <au><snm>Yoo</snm><fnm>JS</fnm></au>
    <au><snm>Hwang</snm><fnm>JH</fnm></au>
    <au><snm>Song</snm><fnm>BK</fnm></au>
    <au><snm>Yoon</snm><fnm>JW</fnm></au>
  </aug>
  <source>International Conference on Information Security Practice and
  Experience</source>
  <pubdate>2019</pubdate>
  <fpage>20</fpage>
  <lpage>-40</lpage>
</bibl>

<bibl id="B12">
  <title><p>Secure Multiparty Linear and Logistic Regression Based on
  Homomorphic Encryption</p></title>
  <aug>
    <au><snm>MLD</snm><fnm>RH</fnm></au>
    <au><snm>Fienberg</snm><fnm>S</fnm></au>
    <au><snm>Nardi</snm><fnm>Y</fnm></au>
  </aug>
  <source>cs.cmu.edu</source>
  <url>http://www.cs.cmu.edu/~rjhall/privacy_day_2011.pdf</url>
  <note>Query date: 2020-06-24 08:59:23</note>
</bibl>

<bibl id="B13">
  <title><p>Privacy-preserving semi-parallel logistic regression training with
  fully homomorphic encryption</p></title>
  <aug>
    <au><snm>Carpov</snm><fnm>S</fnm></au>
    <au><snm>Gama</snm><fnm>N</fnm></au>
    <au><snm>Georgieva</snm><fnm>M</fnm></au>
    <au><snm>Troncoso Pastoriza</snm><fnm>JR</fnm></au>
  </aug>
  <source>BMC Medical Genomics</source>
  <publisher>Springer</publisher>
  <pubdate>2020</pubdate>
  <volume>13</volume>
  <issue>7</issue>
  <fpage>1</fpage>
  <lpage>-10</lpage>
</bibl>

<bibl id="B14">
  <title><p>Secure logistic regression based on homomorphic encryption: Design
  and evaluation</p></title>
  <aug>
    <au><snm>Kim</snm><fnm>M</fnm></au>
    <au><snm>Song</snm><fnm>Y</fnm></au>
    <au><snm>Wang</snm><fnm>S</fnm></au>
    <au><snm>Xia</snm><fnm>Y</fnm></au>
    <au><snm>Jiang</snm><fnm>X</fnm></au>
  </aug>
  <source>JMIR medical informatics</source>
  <publisher>JMIR Publications Inc., Toronto, Canada</publisher>
  <pubdate>2018</pubdate>
  <volume>6</volume>
  <issue>2</issue>
  <fpage>e19</fpage>
</bibl>

<bibl id="B15">
  <title><p>Logistic regression on homomorphic encrypted data at
  scale</p></title>
  <aug>
    <au><snm>Han</snm><fnm>K</fnm></au>
    <au><snm>Hong</snm><fnm>S</fnm></au>
    <au><snm>Cheon</snm><fnm>JH</fnm></au>
    <au><snm>Park</snm><fnm>D</fnm></au>
  </aug>
  <source>Proceedings of the AAAI Conference on Artificial
  Intelligence</source>
  <pubdate>2019</pubdate>
  <volume>33</volume>
  <issue>01</issue>
  <fpage>9466</fpage>
  <lpage>-9471</lpage>
</bibl>

<bibl id="B16">
  <title><p>Efficient Logistic Regression on Large Encrypted Data.</p></title>
  <aug>
    <au><snm>Han</snm><fnm>K</fnm></au>
    <au><snm>Hong</snm><fnm>S</fnm></au>
    <au><snm>Cheon</snm><fnm>JH</fnm></au>
    <au><snm>Park</snm><fnm>D</fnm></au>
  </aug>
  <source>IACR Cryptol. ePrint Arch.</source>
  <publisher>eprint.iacr.org</publisher>
  <pubdate>2018</pubdate>
  <url>https://eprint.iacr.org/2018/662.pdf</url>
  <note>Query date: 2020-06-24 08:59:23</note>
</bibl>

<bibl id="B17">
  <title><p>Privacy-preserving analytics: secure logistic
  regression</p></title>
  <aug>
    <au><snm>Djonatan</snm><fnm>P</fnm></au>
  </aug>
  <pubdate>2019</pubdate>
  <url>https://dr.ntu.edu.sg/handle/10356/77126</url>
  <note>Query date: 2020-06-24 08:59:23</note>
</bibl>

<bibl id="B18">
  <title><p>Privacy-preserving multiparty learning for logistic
  regression</p></title>
  <aug>
    <au><snm>Du</snm><fnm>W</fnm></au>
    <au><snm>Li</snm><fnm>A</fnm></au>
    <au><snm>Li</snm><fnm>Q</fnm></au>
  </aug>
  <source>International Conference on Security and Privacy in Communication
  Systems</source>
  <pubdate>2018</pubdate>
  <fpage>549</fpage>
  <lpage>-568</lpage>
</bibl>

<bibl id="B19">
  <title><p>Privacy-preserving logistic regression</p></title>
  <aug>
    <au><snm>Chaudhuri</snm><fnm>K</fnm></au>
    <au><snm>Monteleoni</snm><fnm>C</fnm></au>
  </aug>
  <source>Advances in neural information processing systems</source>
  <pubdate>2009</pubdate>
  <fpage>289</fpage>
  <lpage>-296</lpage>
</bibl>

<bibl id="B20">
  <title><p>A secure distributed logistic regression protocol for the detection
  of rare adverse drug events</p></title>
  <aug>
    <au><snm>El Emam</snm><fnm>K</fnm></au>
    <au><snm>Samet</snm><fnm>S</fnm></au>
    <au><snm>Arbuckle</snm><fnm>L</fnm></au>
    <au><snm>Tamblyn</snm><fnm>R</fnm></au>
    <au><snm>Earle</snm><fnm>C</fnm></au>
    <au><snm>Kantarcioglu</snm><fnm>M</fnm></au>
  </aug>
  <source>Journal of the American Medical Informatics Association</source>
  <publisher>BMJ Group</publisher>
  <pubdate>2013</pubdate>
  <volume>20</volume>
  <issue>3</issue>
  <fpage>453</fpage>
  <lpage>-461</lpage>
</bibl>

<bibl id="B21">
  <title><p>Secure and differentially private logistic regression for
  horizontally distributed data</p></title>
  <aug>
    <au><snm>Kim</snm><fnm>M</fnm></au>
    <au><snm>Lee</snm><fnm>J</fnm></au>
    <au><snm>Ohno Machado</snm><fnm>L</fnm></au>
    <au><snm>Jiang</snm><fnm>X</fnm></au>
  </aug>
  <source>IEEE Transactions on Information Forensics and Security</source>
  <publisher>IEEE</publisher>
  <pubdate>2019</pubdate>
  <volume>15</volume>
  <fpage>695</fpage>
  <lpage>-710</lpage>
</bibl>

<bibl id="B22">
  <title><p>High-performance secure multi-party computation for data mining
  applications</p></title>
  <aug>
    <au><snm>Bogdanov</snm><fnm>D</fnm></au>
    <au><snm>Niitsoo</snm><fnm>M</fnm></au>
    <au><snm>Toft</snm><fnm>T</fnm></au>
    <au><snm>Willemson</snm><fnm>J</fnm></au>
  </aug>
  <source>International Journal of Information Security</source>
  <publisher>Springer</publisher>
  <pubdate>2012</pubdate>
  <volume>11</volume>
  <issue>6</issue>
  <fpage>403</fpage>
  <lpage>-418</lpage>
</bibl>

<bibl id="B23">
  <title><p>Efficient multiparty protocols using circuit
  randomization</p></title>
  <aug>
    <au><snm>Beaver</snm><fnm>D</fnm></au>
  </aug>
  <source>Annual International Cryptology Conference</source>
  <pubdate>1991</pubdate>
  <fpage>420</fpage>
  <lpage>-432</lpage>
</bibl>

<bibl id="B24">
  <title><p>Achieving both valid and secure logistic regression analysis on
  aggregated data from different private sources</p></title>
  <aug>
    <au><snm>Nardi</snm><fnm>Y</fnm></au>
    <au><snm>Fienberg</snm><fnm>SE</fnm></au>
    <au><snm>Hall</snm><fnm>RJ</fnm></au>
  </aug>
  <source>Journal of Privacy and Confidentiality</source>
  <pubdate>2012</pubdate>
  <volume>4</volume>
  <issue>1</issue>
</bibl>

<bibl id="B25">
  <title><p>Categorical data analysis</p></title>
  <aug>
    <au><snm>Agresti</snm><fnm>A</fnm></au>
  </aug>
  <publisher>John Wiley \& Sons</publisher>
  <pubdate>2003</pubdate>
  <volume>482</volume>
</bibl>

<bibl id="B26">
  <title><p>Secure logistic regression based on homomorphic encryption: Design
  and evaluation</p></title>
  <aug>
    <au><snm>Kim</snm><fnm>M</fnm></au>
    <au><snm>Song</snm><fnm>Y</fnm></au>
    <au><snm>Wang</snm><fnm>S</fnm></au>
    <au><snm>Xia</snm><fnm>Y</fnm></au>
    <au><snm>Jiang</snm><fnm>X</fnm></au>
  </aug>
  <source>JMIR medical informatics</source>
  <publisher>JMIR Publications Inc., Toronto, Canada</publisher>
  <pubdate>2018</pubdate>
  <volume>6</volume>
  <issue>2</issue>
  <fpage>e19</fpage>
</bibl>

<bibl id="B27">
  <title><p>The lower bound method in probit regression</p></title>
  <aug>
    <au><snm>B{\"o}hning</snm><fnm>D</fnm></au>
  </aug>
  <source>Computational statistics \& data analysis</source>
  <publisher>Elsevier</publisher>
  <pubdate>1999</pubdate>
  <volume>30</volume>
  <issue>1</issue>
  <fpage>13</fpage>
  <lpage>-17</lpage>
</bibl>

<bibl id="B28">
  <title><p>{UCI} Machine Learning Repository</p></title>
  <aug>
    <au><snm>Dua</snm><fnm>D</fnm></au>
    <au><snm>Graff</snm><fnm>C</fnm></au>
  </aug>
  <pubdate>2017</pubdate>
  <url>http://archive.ics.uci.edu/ml</url>
</bibl>

<bibl id="B29">
  <title><p>lbw: Low Birth Weight study data</p></title>
  <pubdate>2019</pubdate>
  <url>https://rdrr.io/rforge/LogisticDx/man/lbw.html</url>
</bibl>

<bibl id="B30">
  <title><p>pcs: Prostate Cancer Study data</p></title>
  <pubdate>2019</pubdate>
  <url>https://rdrr.io/rforge/LogisticDx/man/pcs.html</url>
</bibl>

<bibl id="B31">
  <title><p>uis: UMARU IMPACT Study data</p></title>
  <pubdate>2019</pubdate>
  <url>https://rdrr.io/rforge/LogisticDx/man/uis.html</url>
</bibl>

<bibl id="B32">
  <title><p>memory-profiler</p></title>
  <pubdate>2021</pubdate>
  <url>https://pypi.org/project/memory-profiler/</url>
</bibl>

<bibl id="B33">
  <title><p>Sharemind: programmable secure computations with practical
  applications</p></title>
  <aug>
    <au><snm>Bogdanov</snm><fnm>D</fnm></au>
  </aug>
  <source>PhD thesis</source>
  <publisher>Tartu University</publisher>
  <pubdate>2013</pubdate>
</bibl>

</refgrp>
} 
